\pdfoutput=1

\documentclass[11pt]{article}

\usepackage[]{acl}

\usepackage{inconsolata}
\usepackage{times}
\usepackage{latexsym}
\usepackage{amsfonts}
\usepackage{amsmath}
\usepackage{amssymb}
\usepackage{multicol}
\usepackage{multirow}
\usepackage{xspace}
\usepackage{booktabs}
\usepackage{bbding}
\usepackage{array}
\usepackage{threeparttable}
\usepackage{tcolorbox}
\usepackage{tabularx}
\usepackage{enumitem}
\usepackage[linesnumbered,ruled,vlined]{algorithm2e}
\usepackage{xcolor,colortbl}
\usepackage{color}
\usepackage{setspace}
\usepackage{makecell}
\usepackage{listings}
\usepackage{xltabular}

\definecolor{Ocean}{RGB}{129,194,250}

\usepackage{cleveref}
\crefformat{section}{\S#2#1#3}
\crefformat{subsection}{\S#2#1#3}
\crefformat{subsubsection}{\S#2#1#3}
\crefrangeformat{section}{\S\S#3#1#4 to~#5#2#6}
\crefmultiformat{section}{\S\S#2#1#3}{ and~#2#1#3}{, #2#1#3}{ and~#2#1#3}
\crefmultiformat{subsection}{\S\S#2#1#3}{ and~#2#1#3}{, #2#1#3}{ and~#2#1#3}
\Crefformat{figure}{#2Fig.~#1#3}
\Crefmultiformat{figure}{Figs.~#2#1#3}{ and~#2#1#3}{, #2#1#3}{ and~#2#1#3}
\Crefformat{table}{#2Tab.~#1#3}
\Crefmultiformat{table}{Tabs.~#2#1#3}{ and~#2#1#3}{, #2#1#3}{ and~#2#1#3}
\Crefformat{appendix}{Appx.~\S#2#1#3}
\crefmultiformat{appendix}{Appx.~\S#2#1#3}{ and~#2#1#3}{, #2#1#3}{ and~#2#1#3}
\crefformat{algorithm}{Alg.~#2#1#3}
\Crefformat{equation}{Eq.~#2#1#3}

\newcommand{\xhdr}[1]{\vspace{0.3em}\noindent{{\bf #1.}}}
\newcommand{\modelname}{\textsc{Zooter}\xspace}
\newcommand{\datasetname}{\textsc{DivInstruct}\xspace}

\usepackage[T1]{fontenc}

\usepackage[utf8]{inputenc}

\usepackage{microtype}
\usepackage{amsmath}
\usepackage{amsthm}
\title{Routing to the Expert: Efficient Reward-guided Ensemble of Large Language Models}

\author{
Keming Lu, Hongyi Yuan\thanks{$^*$Work done during internship at Alibaba Inc.} , Runji Lin$^*$\\ \bf{Junyang Lin, Zheng Yuan, Chang Zhou, Jingren Zhou}
\\
Alibaba Inc. \\
\texttt{\{lukeming.lkm,yuanhongyi.yhy,linrunji.lrj\}@alibaba-inc.com}\\
\texttt{\{junyang.ljy,yuanzheng.yuanzhen\}@alibaba-inc.com}\\
\texttt{\{ericzhou.zc,jingren.zhou\}@alibaba-inc.com}\\
}

\begin{document}
\maketitle
\begin{abstract}
The complementary potential of Large Language Models~(LLM) assumes off-the-shelf LLMs have heterogeneous expertise in a wide range of domains and tasks so that an ensemble of LLMs can achieve consistently better performance.
Existing ensemble methods for LLMs mainly focus on reward model ranking of outputs, leading to significant computation overhead.
To combat this issue, we revisit the complementary potential of LLMs and further elaborate it by mining latent expertise with off-the-shelf reward models.
We propose \modelname, a reward-guided routing method distilling rewards on training queries to train a routing function, which can precisely distribute each query to the LLM with expertise about it.
We also integrate a tag-based label enhancement to mitigate noise from uncertainty when using rewards as silver supervision.
\modelname shows computation efficiency in inference as it only introduces minor computation overhead of a routing function compared with reward model ranking methods.
We evaluate \modelname on a comprehensive benchmark collection with 26 subsets on different domains and tasks.
\modelname outperforms the best single model on average and ranks first on 44\% of tasks, even surpassing multiple reward model ranking methods.
\footnote{Work in progress.}
\end{abstract}

\section{Introduction}
Large Language Models~(LLMs) aligned with human preference rapidly emerge and are almost daily released~\cite{touvron2023llama,llama2,anil2023palm,bai2023qwen}.
These off-the-shelf LLMs are further finetuned or aligned with human preference to be generalists~\cite{xu2023wizardlm,llama2,touvron2023llama} or specialists~\cite{rft,luo2023wizardmath,luo2023wizardcoder,roziere2023code} for solving versatile tasks.
It is worth noticing that LLMs are pretrained and aligned with various data, leading to diverse strengths and weaknesses in versatile downstream tasks~\cite{jiang2023llm}.
Therefore, the ensemble of LLMs harnesses the complementary potential among them and may achieve better performance than a single best-on-average model across diverse tasks.

\begin{figure}[t]
    \centering
    \includegraphics[width=\linewidth]{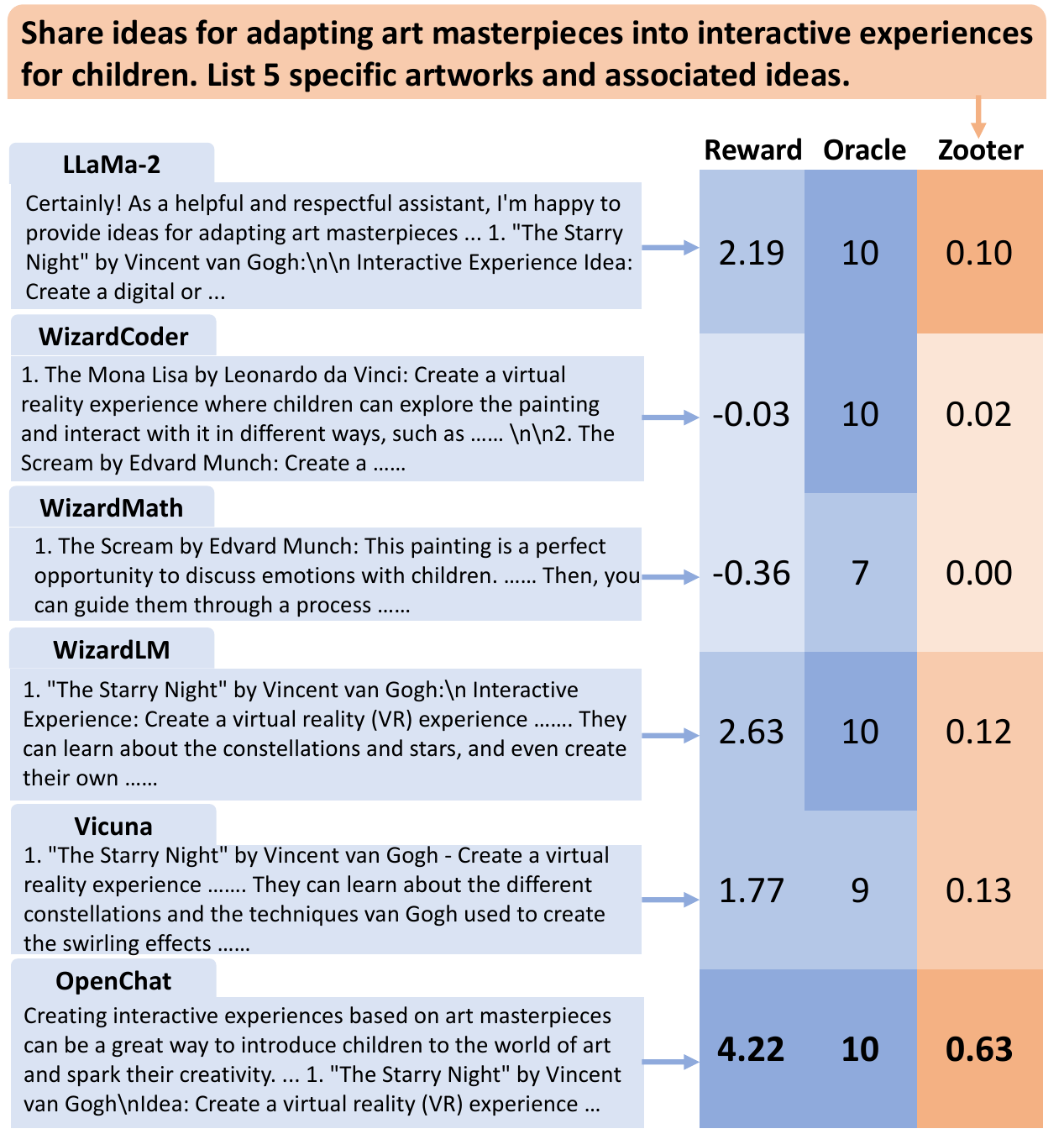}
    \caption{
    An example of the large language model ensemble.
    Reward model ranking marked in blue needs to generate responses from all models while \modelname routers the given query to the best model and only infers one model.
    This case is collected from the MT-Bench benchmark and we also present oracle judgements of each response.
    }
    \label{fig:teaser}
    \vspace{-2em}
\end{figure}

One of the key challenges in the LLM ensemble is computation efficiency due to the large parameter size of existing LLMs.
Previous research~\cite{jiang2023llm,shnitzer2023large} provides solid methods to merge generation outputs of LLMs as an ensemble.
Such methods require tremendous inference cost that makes it unscalable and thus not competitive to the best-on-average model under low-resource scenarios.
To efficiently assemble off-the-shelf LLMs, we first dive deeper into the considerably straightforward but still understudied assumption: \textit{Off-the-shelf aligned LLMs, even for those aligned as ``generalists'', have heterogeneous expertise in a wide range of domains and topics}.
However, analyzing the expertise of an LLM is also challenged as the latent expertise of LLMs is highly related to the pretrained and alignment data, which is very vague and inaccessible even for popular open-source LLMs such as \textsc{Llama-2-Chat}~\cite{llama2} and \textsc{WizardLM}~\cite{xu2023wizardlm}.

If this assumption strongly holds, off-the-shelf LLMs can be assembled efficiently by assigning queries to the model that is proficient in the query without additional inference costs on each model.
Such an efficient routing strategy only requires inference cost for a single model for each query and the overhead cost of a much smaller query router.
However, probing the detailed expertise of off-the-shelf LLMs and generating supervision for training routers also require annotations.
Developing a data-efficient training method for routing queries is significantly understudied.

To combat these issues, we propose \modelname, a reward-guided query routing method for efficiently assembling off-the-shelf LLMs.
\modelname obtains and enhances silver supervision from existing reward models~(RM) for query router training and distributes queries in advance to ``expertise''.
As shown in \Cref{fig:teaser}, the reward distribution implies the oracle judgments and reveals a latent expertise between LLMs.
And \modelname captures the expertise from reward distributions and provides query distribution during inference.
Specifically, we first conduct a comprehensive study involving four groups of benchmarks across more than 26 subsets in various domains and tasks.
We investigate six widely used open-source LLMs and show the complementary potential of such wide-range downstream tasks by aggregating them via reward model ranking.
We then collect a diverse training query set and distill rewards of model expertise as indirect supervision for training an LLM router and develop tag-based label enhancement to overcome the shortage of such silver labels from reward models further.
With comprehensive experiments, we show \modelname can benefit from RM silver supervision to learn the latent expertise among LLMs and conduct efficient routing for the model ensemble.
Our contributions are mainly three-fold: 
\begin{itemize}[leftmargin=1em]
    \setlength\itemsep{0em}
    \item  We revisit the complementary potential of open-source LLMs, which proves the effectiveness of LLM ensemble, and show rewards from off-the-shelf RMs can be silver supervision for model expertise.
    \item We propose \modelname, an efficient reward-guided routing method distilling rewards from off-the-shelf reward model for probing model expertise. Then, we develop a tag-based label enhancement to mitigate noise from the uncertainty of reward models.
    \item We comprehensively evaluate ensemble methods, including reward model ranking and \modelname on four groups of benchmarks with 26 subsets on different tasks and domains. Our evaluation shows \modelname can effectively assemble LLMs and even outperforms reward model ranking methods with significantly fewer computation overhead.
\end{itemize}


\section{Related Works}
\xhdr{Instruction Tuning and Alignment}
Instruction tuning~\cite{longpre2023flan} helps LLMs to follow versatile instructions, which is widely adopted to align LLMs with human preference~\cite{vicuna2023,xu2023wizardlm,bai2023qwen}.
In this work, we focus on assembling aligned LLMs, such as Llama-2-Chat~\cite{llama2}, WizardLM~\cite{xu2023wizardlm}, Vicuna~\cite{vicuna2023}, and so on.
And we evaluate them on a wide range of alignment evaluation tasks.

\xhdr{Large Language Model Ensemble}
The ensemble of LLMs is an emerging topic due to the explosion of open-source LLMs.
LLM ensemble aims to merge off-the-shelf LLMs to achieve consistently better performance across diverse downstream tasks.
Few works explore the complementary potential assumption of LLMs and how to assemble LLMs with it.
\citet{jiang2023llm} presents an ensembling framework consisting of a pair ranker and a generation fuser.
\citet{chen2023frugalgpt} sequentially infers off-the-shelf LLMs and stops until the response meets a sufficient quality.
\citet{wang2023fusing} proposes a fusing-of-experts problem that fuses outputs of expert models with complementary knowledge of the data distribution and formulates it as supervised learning.
\citet{shnitzer2023large} show the utility and limitations of learning model routers from various benchmark datasets.
Although these works all focus on reward ranking or routing strategies to assemble LLMs, \modelname distinguishes from these concurrent works in two aspects.
First, our concurrent works require output generations or the forward process to get prompt representations of all candidates, leading to significant computation overhead.
\modelname infers model expertise by distilling rewards on a predefined training query set to avoid such inference overhead.
Then, all these works are developed and evaluated on a set of benchmarks, while \modelname can be developed with only queries without golden responses, and \modelname aims for more diverse alignment tasks.
Therefore, \modelname stands out for its efficiency in data and computation.
We also evaluate \modelname on more diverse alignment tasks to comprehensively examine the complementary potential of LLMs.

\xhdr{Reward Model Guided Generation}
Reward models in the context of large language models are commonly used to improve alignment performance by reinforcement learning \citep{ppo,instructgpt} or preference learning \citep{yuan2023rrhf,dpo,pro}.
Reward models can also improve the performance during the generation phase.
The math reasoning ability of language models can be improved by using reward models ranking multiple generated reasoning paths \citep{cobbe2021training,uesato2022solving,lightman2023lets}.
\citet{liu2023dont} uses reward models to formulate reward-guided decoding.
Inspired by these successful applications of reward models in alignment, \modelname also takes advantage of off-the-shelf reward models to investigate the latent expertise of LLMs.

\begin{figure*}[t]
    \centering
    \includegraphics[width=\linewidth]{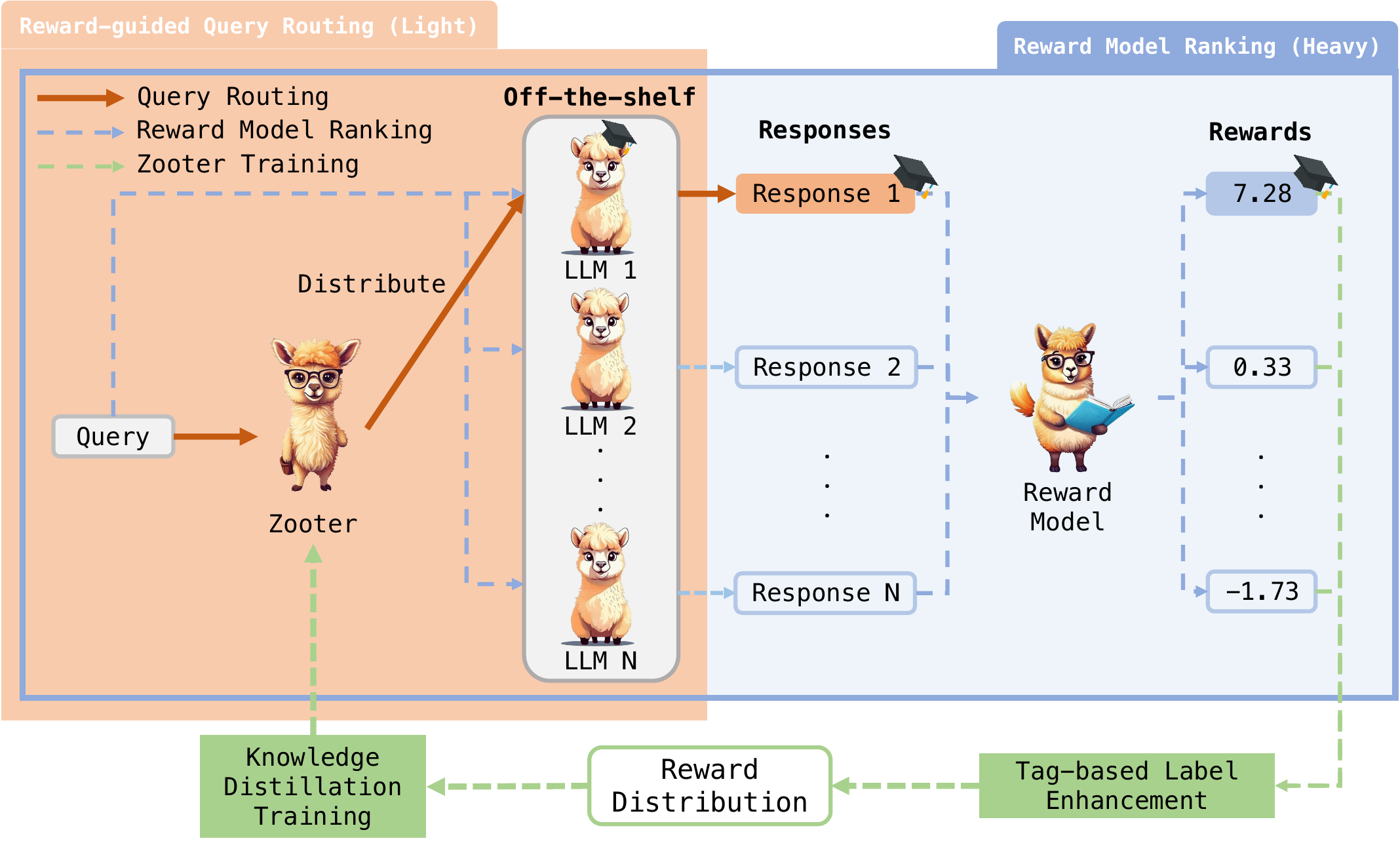}
    \caption{
    Overview of \modelname.
    \modelname aims to assemble a set of off-the-shelf LLMs by first conducting a reward model ranking on a diverse training set to obtain supervision of model expertise, highlighted in blue in the figure.
    Instruction tags are then used to mitigate the uncertainty in reward estimation.
    \modelname uses the normalized rewards as supervision to train a routing function by knowledge distillation.
    The training circle is marked in green, and the inference is marked in orange.
    \modelname is much lighter in computation as it routes the query to the corresponding expert LLM during inference time, while reward model ranking has to generate outputs for all candidates.
    }
    \label{fig:main}
\end{figure*}

\section{Methods}
We first revisit the complementary potential of LLMs~(\Cref{sec:complementary_potential}) and then introduce \modelname as an efficient LLM ensemble method~(\Cref{sec:zooter}).

\subsection{Complementary Potential of LLMs}\label{sec:complementary_potential}

In this section, we present the preliminaries about the assumption: \textit{Off-the-shelf
aligned LLMs have heterogeneous expertise in a wide
range of domains and topics}.
We also briefly introduce two LLM ensemble strategies, reward model ranking, and query routing.

\xhdr{Complementary Potential Assumption} Considering a set of LLMs denoted as $\mathcal{M}=\{m_i|i\in \mathcal{Z}^+\}$ and a set of downstream queries denoted as $\mathcal{Q}=\{q_i|i\in \mathcal{Z}^+\}$, we assume that for each LLM $m_i$ in $\mathcal{M}$, there exists a non-empty query subset $\mathcal{Q}_{m_i}$ such that the LLM can achieve uniformly better performance than other LLMs in $\mathcal{M}$ for any query $q_j\in\mathcal{Q}_{m_i}$, which is 
$m_i = \operatorname{argmax}_{m\in \mathcal{M}} P\left(q_j, m(q_j)\right)$.
$P$ can be any preference or metric for performance assessment.
In this work, we further enhance this assumption and aim to show that the complementary between LLMs reveals their expertise in different domains and tasks, so that we can categorize queries and choose the best LLM for each category.

\xhdr{Reward Model Ranking} Reward model ranking~(RMR) leverages the complementary potential to ensemble LLMs and achieve surpass performance. RMR tries to find a reward function $\hat{P}$ to estimate the oracle preference $P$ so that we can obtain the best model for each query~\cite{jiang2023llm}.
However, RMR infers all candidate models to get outputs and then rank them with a reward function, introducing a large computation overhead.

\xhdr{Query Routing} Query routing mitigates efficiency concerns in the LLM ensemble, especially compared with existing RMR methods.
In general, query routing tries to find a routing function $\mathcal{Z}(q, m_i)$ with respect to $q_j\in \mathcal{Q}$ exists, so that $m_i = \operatorname{argmax}_{m\in \mathcal{M}} \mathcal{Z}\left(q_j, m\right)$.
The routing function distributes queries based on themselves without generating outputs.
If the complementary potential of LLMs holds, the routing function predicts the probability that a query $q$ belongs to the expertise of an LLM $Q_{m}$.

\subsection{Zooter}\label{sec:zooter}

In this section, we propose \textsc{Zooter}, a reward-guided query routing method for efficiently assembling large language models.
\modelname learns from the reward model ranking to interpret the latent expertise of each model.
So, as shown in \Cref{fig:main}, \modelname first infers all candidate LLMs on a training set containing diverse queries to generate responses.
Then, all responses will be rewarded by an off-the-shelf reward model providing scalar rewards, marked in blue dash lines in \Cref{fig:main}.
The rewards are first enhanced by a tag-based prior for smoothing and denoising.
The normalized reward distribution is then used as supervision in the knowledge distillation training of the routing function, shown in the green dash lines in \Cref{fig:main}.
During inference, the routing function categorizes the input query to an LLM with the strongest expertise potential in this query, and the LLM will generate an expert response.
By training such a routing function, \modelname achieves a much more efficient ensemble as it only needs to infer one expert LLM, plus a small computation overhead of the routing function.
In this section, we introduce the two key components along with the design motivations.

\xhdr{Reward Distillation}
As we discussed in \Cref{sec:complementary_potential}, query routing aims to find a routing function predicting the probability that a query $q$ belongs to the expertise of an LLM $Q_{m}$, where $Q_{m}$ is a set of queries that an LLM $m$ consistently achieves maximum preference among all candidates.
Recalling the reward model ranking, we notice the estimated preferences $\hat{P}(q, m_i(q))$, i.e., reward, can be interpreted as the relative advantages of an LLM $m_i$ among all candidates on the query $q$.
Therefore, the normalized reward can be used as a silver supervision for the routing function:
\begin{equation*}
\begin{aligned}
    \mathcal{Z}(q)_i &= P(q\in Q_{m_i})\\ &:= \frac{\exp(\hat{P}(q, m_i(q)))}{\sum_{m_i \in \mathcal{M}}\exp(\hat{P}(q, m_i(q)))},
\end{aligned}
\end{equation*}
as the higher advantages inherently present the expertise of an LLM on a query compared with its counterparts.

To estimate the expertise of each model and train the routing function, we need to apply the reward preference ranking on a diverse training set $\hat{Q}$.
We first infer all candidate models on each query $\hat{q}\in \hat{Q}$, and then assign rewards by an off-the-shelf reward model to obtain a scalar reward for each query and model \begin{equation*}
    \mathbf{r}_{i}=\{\hat{P}(\hat{q}_i, m_j(\hat{q}_i))\}_{j=1}^{\vert \mathcal{M} \vert}\;,\;i=1,\ldots,\vert \hat{Q} \vert.
\end{equation*}.
Then, we train the router function $\mathcal{Z}$ on the training set by knowledge distillation with a Kullback-Leibler divergence as the loss function:
\begin{equation*}
    \mathcal{L}(q_i, \mathbf{r}_i) = \textrm{KL}(\mathcal{Z}(q_i), \textrm{softmax}(\mathbf{r}_i)).
\end{equation*}
\modelname is a data-efficient and low-resource method as the training set $\hat{Q}$ only contains queries without annotations of responses.
However, queries in the training set are expected to be as diverse as possible to maximize the generalization abilities of the routing function.
The distillation process helps \modelname to learn the latent expertise of each model.
So, we can mitigate the computation cost by only judging whether a query belongs to the expertise set with our routing function during inference.

\xhdr{Tag-based Label Enhancement}
Although reward distillation provides a feasible way for routing functions to leverage silver supervision from reward model ranking, the language reward model provides rewards with uncertainty, introducing certain noises~\cite{gleave2022uncertainty}.
We first empirically analyze this uncertainty in \Cref{sec:analysis}.
Existing off-the-shelf reward models will all involve noises in terms of uncertainty, as shown in \Cref{fig:rm_uncertainty}.
Therefore, we leverage instruction tagging to enhance rewards on the training queries further.
The tag-based label enhancement we proposed is similar to the widely used label smoothing techniques and proven effective in knowledge distillation~\cite{yuan2020revisiting}.
Specifically, we first tag each query $\hat{q}_i \in \hat{Q}$ with a local tagger $\mathcal{T}(\cdot)$ to obtain a set of tags $\mathcal{T}(q_i)$.
Then, we aggregate all rewards on queries with the same tags for the tag-wise rewards as follows:
\begin{equation*}
    \begin{aligned}
    Q_t &= \{q_i|t\in \mathcal{T}(q_i), i=1,\ldots,\vert\hat{Q}\vert\} \\
    \mathbf{r}_t &= \frac{1}{\vert Q_t \vert}\sum_{i\in Q_t} \mathbf{r_i}
    \end{aligned}
\end{equation*}
Then, we enhance rewards for each query with tag-wise rewards by a linear combination:
\begin{equation*}
    \mathbf{r}^*_i = \beta \mathbf{r}_i + (1-\beta)\mathbf{r}_t\;;\;t=\mathcal{T}(q_i),i=1,\ldots,\vert\hat{Q}\vert
\end{equation*}
,where $\beta$ is a hyper-parameter for the trade-off between coarse-grained tag-wise rewards and fine-grained sample-level rewards.
Then, we replace original rewards in the KL divergence loss training with tag-based enhanced rewards $\mathbf{r}^*$ during routing function training.

\section{Experiments}
In this section, we report experimental setup~(\Cref{sec:experimental_setup}), main results~(\Cref{sec:results}), and analysis about \modelname~(\Cref{sec:analysis}).

\begin{table*}[t]
    \centering
    \small
    \setlength{\tabcolsep}{0.5mm}{
    \begin{tabular}{lcc|cccccccc|cc}
    \toprule
    \multirow{2}{*}{\textbf{Model}} & \multicolumn{2}{c}{\textbf{\#Param}} & \multicolumn{2}{c}{\textbf{AlpacaEval~(5)}} & \multicolumn{2}{c}{\textbf{FLASK~(10)}} & \multicolumn{2}{c}{\textbf{MT-Bench~(8)}} & \multicolumn{2}{c}{\textbf{Benchmarks~(3)}} & \multicolumn{2}{c}{\textbf{All~(26)}} \\
    & Ranker & Infer & Avg. & MTR & Avg. & MTR & Avg. & MTR & Avg. & MTR & MTR & \% Uplift \\
    \midrule
    \multicolumn{12}{c}{\textit{Routing Candidates}}\\
    \midrule
	\textsc{WizardCoder} & $--$ & 13B & \cellcolor{Ocean!0}0.42 & \cellcolor{Ocean!0}5.6 & \cellcolor{Ocean!0}3.12 & \cellcolor{Ocean!0}5.2 & \cellcolor{Ocean!0}4.44 & \cellcolor{Ocean!0}5.38 & \cellcolor{Ocean!0}30.9 & \cellcolor{Ocean!0}4.33 & \cellcolor{Ocean!0}5.3 & \cellcolor{Ocean!3}0.06 \\
	\textsc{WizardLM} & $--$ & 13B & \cellcolor{Ocean!83}0.89 & \cellcolor{Ocean!79}2.0 & \cellcolor{Ocean!53}3.89 & \cellcolor{Ocean!81}1.8 & \cellcolor{Ocean!59}7.15 & \cellcolor{Ocean!78}2.0 & \cellcolor{Ocean!23}44.2 & \cellcolor{Ocean!70}2.0 & \cellcolor{Ocean!81}1.83 & \cellcolor{Ocean!22}0.25 \\
	\textsc{WizardMath} & $--$ & 13B & \cellcolor{Ocean!8}0.47 & \cellcolor{Ocean!14}5.0 & \cellcolor{Ocean!11}3.28 & \cellcolor{Ocean!5}5.0 & \cellcolor{Ocean!28}5.73 & \cellcolor{Ocean!23}4.38 & \cellcolor{Ocean!6}34.8 & \cellcolor{Ocean!10}4.0 & \cellcolor{Ocean!17}4.6 & \cellcolor{Ocean!0}0.03 \\
	\textsc{Llama-2-chat} & $--$ & 13B & \cellcolor{Ocean!87}0.91 & \cellcolor{Ocean!87}1.6 & \cellcolor{Ocean!52}3.88 & \cellcolor{Ocean!89}1.5 & \cellcolor{Ocean!50}6.72 & \cellcolor{Ocean!58}2.88 & \cellcolor{Ocean!2}32.3 & \cellcolor{Ocean!20}3.67 & \cellcolor{Ocean!72}2.23 & \cellcolor{Ocean!28}0.31 \\
	\textsc{OpenChat} & $--$ & 13B & \cellcolor{Ocean!83}0.89 & \cellcolor{Ocean!74}2.2 & \cellcolor{Ocean!46}3.79 & \cellcolor{Ocean!50}3.1 & \cellcolor{Ocean!58}7.12 & \cellcolor{Ocean!78}2.0 & \cellcolor{Ocean!0}31.2 & \cellcolor{Ocean!31}3.33 & \cellcolor{Ocean!62}2.67 & \cellcolor{Ocean!16}0.19 \\
	\textsc{Vicuna} & $--$ & 13B & \cellcolor{Ocean!67}0.8 & \cellcolor{Ocean!40}3.8 & \cellcolor{Ocean!40}3.7 & \cellcolor{Ocean!41}3.5 & \cellcolor{Ocean!47}6.58 & \cellcolor{Ocean!49}3.25 & \cellcolor{Ocean!4}33.6 & \cellcolor{Ocean!50}2.67 & \cellcolor{Ocean!45}3.4 & \cellcolor{Ocean!3}0.06 \\
\midrule
	\textsc{BMA} & $--$ & 13B & \cellcolor{Ocean!87}0.91 & \cellcolor{Ocean!87}1.6 & \cellcolor{Ocean!52}3.88 & \cellcolor{Ocean!89}1.5 & \cellcolor{Ocean!50}6.72 & \cellcolor{Ocean!58}2.88 & \cellcolor{Ocean!2}32.3 & \cellcolor{Ocean!20}3.67 & \cellcolor{Ocean!72}2.23 & \cellcolor{Ocean!28}0.31 \\
\midrule
\multicolumn{12}{c}{\modelname}\\
\midrule
	Ours & 86M & 13B & \cellcolor{Ocean!91}0.93 & \cellcolor{Ocean!97}1.17 & \cellcolor{Ocean!53}3.89 & \cellcolor{Ocean!81}1.82 & \cellcolor{Ocean!58}7.11 & \cellcolor{Ocean!70}2.33 & \cellcolor{Ocean!5}34.2 & \cellcolor{Ocean!40}3.0 & \cellcolor{Ocean!79}1.94 & \cellcolor{Ocean!42}0.44 \\
\midrule
\multicolumn{12}{c}{\textit{Reward Model Ranking~(RMR)}}\\
\midrule
	\textsc{w/ OAssistRM} & 300M & 6$\times$13B & \cellcolor{Ocean!66}0.79 & \cellcolor{Ocean!35}4.0 & \cellcolor{Ocean!43}3.75 & \cellcolor{Ocean!35}3.73 & \cellcolor{Ocean!47}6.59 & \cellcolor{Ocean!50}3.22 & \cellcolor{Ocean!7}35.1 & \cellcolor{Ocean!33}3.25 & \cellcolor{Ocean!44}3.42 & \cellcolor{Ocean!16}0.19 \\
	\textsc{w/ LLM-Blender} & 300M & 6$\times$13B & \cellcolor{Ocean!73}0.83 & \cellcolor{Ocean!42}3.67 & \cellcolor{Ocean!45}3.77 & \cellcolor{Ocean!44}3.36 & \cellcolor{Ocean!38}6.21 & \cellcolor{Ocean!32}4.0 & \cellcolor{Ocean!9}36.4 & \cellcolor{Ocean!48}2.75 & \cellcolor{Ocean!45}3.39 & \cellcolor{Ocean!14}0.17 \\
	\textsc{w/ Auto-J} & 13B & 6$\times$13B & \cellcolor{Ocean!83}0.89 & \cellcolor{Ocean!64}2.67 & \cellcolor{Ocean!55}3.92 & \cellcolor{Ocean!85}1.64 & \cellcolor{Ocean!56}7.03 & \cellcolor{Ocean!73}2.22 & \cellcolor{Ocean!2}32.2 & \cellcolor{Ocean!25}3.5 & \cellcolor{Ocean!71}2.25 & \cellcolor{Ocean!40}0.42 \\
	\textsc{w/ UltraRM} & 13B & 6$\times$13B & \cellcolor{Ocean!89}0.92 & \cellcolor{Ocean!97}1.17 & \cellcolor{Ocean!65}4.06 & \cellcolor{Ocean!100}1.0 & \cellcolor{Ocean!60}7.18 & \cellcolor{Ocean!80}1.89 & \cellcolor{Ocean!16}40.1 & \cellcolor{Ocean!33}3.25 & \cellcolor{Ocean!88}1.53 & \cellcolor{Ocean!71}0.72 \\
	\textsc{w/ QwenRM} & 7B & 6$\times$13B & \cellcolor{Ocean!89}0.92 & \cellcolor{Ocean!93}1.33 & \cellcolor{Ocean!63}4.04 & \cellcolor{Ocean!100}1.0 & \cellcolor{Ocean!61}7.26 & \cellcolor{Ocean!75}2.11 & \cellcolor{Ocean!13}38.6 & \cellcolor{Ocean!40}3.0 & \cellcolor{Ocean!87}1.58 & \cellcolor{Ocean!65}0.67 \\
	\textsc{w/ Oracle} & $--$ & 6$\times$13B & \cellcolor{Ocean!100}0.98 & \cellcolor{Ocean!100}1.0 & \cellcolor{Ocean!100}4.56 & \cellcolor{Ocean!100}1.0 & \cellcolor{Ocean!83}8.25 & \cellcolor{Ocean!100}1.0 & \cellcolor{Ocean!77}75.3 & \cellcolor{Ocean!100}1.0 & \cellcolor{Ocean!100}1.0 & \cellcolor{Ocean!100}1.0 \\
\midrule
\multicolumn{12}{c}{\textit{Proprietary Models}}\\
\midrule
	GPT-3.5-turbo & $--$ & $--$ & \cellcolor{Ocean!83}0.89 & \cellcolor{Ocean!64}2.67 & \cellcolor{Ocean!65}4.06 & \cellcolor{Ocean!79}1.91 & \cellcolor{Ocean!76}7.94 & \cellcolor{Ocean!83}1.78 & \cellcolor{Ocean!73}73.0 & \cellcolor{Ocean!100}1.0 & \cellcolor{Ocean!82}1.78 & \cellcolor{Ocean!59}0.61 \\
	GPT-4 & $--$ & $--$ & \cellcolor{Ocean!92}0.94 & \cellcolor{Ocean!100}1.0 & \cellcolor{Ocean!86}4.37 & \cellcolor{Ocean!100}1.0 & \cellcolor{Ocean!100}8.99 & \cellcolor{Ocean!100}1.0 & \cellcolor{Ocean!100}88.3 & \cellcolor{Ocean!100}1.0 & \cellcolor{Ocean!100}1.0 & \cellcolor{Ocean!100}1.0 \\
    \bottomrule
    \end{tabular}}
    \caption{
    Main results of both \modelname and reward model ranking.
    We report performance across four groups of benchmarks and report the number of subsets beside the name of benchmarks.
    We also report the parameters of ranker and total inference models for both candidates and ensemble methods.
    MTR denotes the mean task rate, and \%Uplift denotes the rate of uplift.
    The average scores and uplift rate are as higher as better while MTR is as lower as better.
    We mark better scores in darker blue for better visualization and easier interpretation.
    }
    \label{tab:main_result}
\end{table*}

\subsection{Experimental Setup}\label{sec:experimental_setup}

\xhdr{Candidate LLMs}
We select six \textsc{LLaMa}-based LLMs of the same 13B size as the candidate LLMs for query routing.
(a) \textbf{WizardLM}~\cite{xu2023wizardlm} is aligned with queries and responses augmented by \textsc{EvolInstruct},
(b) \textbf{WizardCoder}~\cite{luo2023wizardcoder} is a coding expert LLM using the same techniques as WizardLM,
(c) \textbf{WizardMath}~\cite{luo2023wizardmath} is a math expert LLM aligned with query augmentation, ChatGPT rewards and PPO optimization,
(d) \textbf{Vicuna}~\cite{vicuna2023} is aligned on tremendous conversations between users and proprietary chatbots,
(e) \textbf{OpenChat}~\cite{openchat} is aligned with a selected set of ShareGPT with additional training strategies,
(f) \textbf{Llama-2-Chat}~\cite{llama2} is first aligned by supervised fine-tuning and then multi-turn rejection sampling.
Both baselines and \modelname are experimented and evaluated based on these six candidates.

\xhdr{Training Datasets}
We create a diverse mix instruction dataset from the open-source data to maximize the generalization abilities of \modelname.
We first collect and tag open-source data from 13 datasets with a local tagger developed by \citet{lu2023instag}.
For trustworthy evaluation results, we decontaminate all samples containing queries that have a 6-gram overlap with any samples in our benchmarks described below to avoid data leakage.
Then, we randomly select ten samples for each unique tag to form a diverse mix instruction dataset \datasetname with 47,986 instructions and samples across 6,270 different tags.
Detailed statistics of \datasetname is in \Cref{app:dataset}.

\xhdr{Benchmarks}
We actively involve four sets of benchmarks to evaluate \modelname on various downstream tasks comprehensively.
We first include three widely-used alignment benchmarks with GPT-4 judge:
\begin{itemize}[leftmargin=1em]
    \setlength\itemsep{0em}
    \item \textbf{AlpcaEval}~\cite{alpaca_eval} consists of 5 subsets from the koala, vicuna, and others evaluation sets. It contains 805 samples in total.
    \item \textbf{FLASK}~\cite{ye2023flask} is a fine-grained evaluation for alignment. We evaluate 10 domains in FLASK and report the average score across all domains as a final score.
    \item \textbf{MT-Bench}~\cite{vicuna2023} is a multi-turn evaluation across eight aspects, including mathematics and coding. We only train and route with the first-turn query but evaluate in the multi-turn manner as the original recipe.
\end{itemize}
However, as reported by \citet{wang2023far}, GPT-4 judgments may have bias and significant disagreement with humans.
Therefore, we also include a group of benchmarks consisting of  MMLU~\cite{mmlu}, GSM8K~\cite{cobbe2021training}, and HumanEval~\cite{humaneval}.

\xhdr{Metrics}
Comparing ensemble models on various benchmarks is challenging as the scale of scores is different on each benchmark.
To combat this issue, we do not only report the scores on each benchmark but also the mean task rank~(MTR).
All benchmarks we evaluate have multiple subsets, we define MTR as the rank of the evaluated model among all baselines average on all subsets.
MTR is only about the rank among baselines so it can be easily adopted across benchmarks that have different score scales.
Similarly, we also propose an uplift rate, denoting the rate of subsets that the evaluated model achieves the best performance of benchmarks.
We report these two metrics on a total of 26 evaluation subsets in all benchmarks.
Lower MTR and higher uplift rates show the evaluated model has consistently higher performance among versatile downstream tasks.

\xhdr{Baselines}
We also compare \modelname with existing reward model ranking~(RMR) methods.
We set up RMR baselines with the latest rewards models, including \textsc{OAssistRM}, \textsc{AUTO-J}~\cite{li2023generative}, \textsc{UltraRM}~\cite{cui2023ultrafeedback}, \textsc{QwenRM}~\cite{bai2023qwen}, and an Oracle ranking for reference.
We also consider the pair ranking in LLM-Blender~\cite{jiang2023llm} as one of the RMR methods.
Besides, we also report the performance of proprietary models across our benchmark collections for reference, including GPT-3.5-turbo and GPT-4.

\xhdr{Configurations}
We train our routing function from \textit{mdeberta-v3-base}.
And we use QwenRM to generate rewards on training queries as supervision for our routing function, as it achieves the best performance in reward model ranking with considerably smaller model parameters described in \Cref{sec:results}.
And we run all training and inference on 8 A100 GPUs.
We infer and evaluate all benchmarks with corresponding configurations and GPT-4 settings.
We use greedy decoding for MMLU, GSM8K, and HumanEval.

\subsection{Results}\label{sec:results}
We present the main results in \Cref{tab:main_result}.
We report the performance of six routing candidates across our benchmarks, and the best model on average~(BMA) is \textsc{LLaMa-2-Chat}.
And we report \modelname with $\beta=0.3$ in tag-based label enhancement.
We further analyze the results in the following two aspects:

\xhdr{Complementary Potential}
We evaluate the ensemble with reward model ranking~(RMR) on five different off-the-shelf reward models.
RMR with UltraRM achieves the best performance in MTR and uplift rate on the aggregation of all benchmarks, which ranks at 1.53 and achieves the best model across 72\% subtasks.
RMR with QwenRM achieves the second best and has similar performance with UltraRM with smaller parameter sizes, followed by RMR with Auto-J, LLM-Blender, and OAssistRM.
RMR with QwenRM, UltraRM, and Auto-J outperform that of BMA, showing the effectiveness of RMR.
Furthermore, we also calculate the score of RMR with an Oracle ranker, which consistently outperforms all candidates and even outperforms GPT-4 on AlpacaEval and FLASK.
Such results provide solid evidence for the complementary potential of off-the-shelf LLMs and also support the key motivation behind \modelname, i.e., using rewards from off-the-shelf reward models as silver supervision for the routing function training.
However, we notice RMR fails on benchmarks, such as MMLU, GSM8K, and HumanEval, showing that precisely judging knowledge, mathematics, and coding problems are still challenging for existing RMs.

\xhdr{Zooter Performance}
We then compare the performance of \modelname with that of BMA and RMR.
\modelname outperforms BMA on AlpacaEval, MT-Bench, and Benchmarks, and achieves similar performance on FLASK.
The most significant improvement is witnessed on MT-Bench, where the performance of \modelname is higher than that of BMA by 0.39.
In general, \modelname achieves top-1 on 44\% subtasks while BMA is only on 31\%.
With the evidence above, \modelname successfully utilizes the complementary potential between LLMs to achieve the best performance more consistently over our benchmarks, with computation overhead from only 86M ranker.
At the same time, \modelname outperforms RMR with OAssistRM, LLM-Blender, and Auto-J, by significantly less computation overhead.
However, though \modelname outperforms RMR with QwenRM on AlpacaEval, there are still obvious gaps between \modelname and RMR with QwenRM in general.

\subsection{Analysis}\label{sec:analysis}
We provide further analysis on how RM uncertainty may influence the training of \modelname.

\begin{figure}[t]
    \centering
    \includegraphics[width=\linewidth]{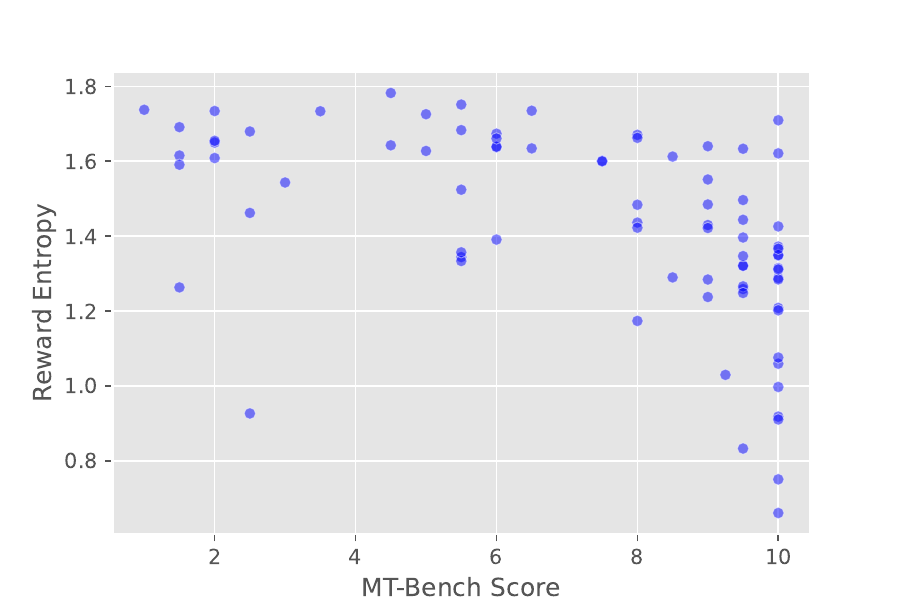}
    \caption{Analysis between reward entropy and scores of reward preference ranking on MT-bench.}
    \label{fig:rm_uncertainty}
\end{figure}

\xhdr{RM Uncertainty}
As presented in the previous research, RM may have uncertainty on its scalar rewards, which may introduce noise in the routing training since we use RM scores as silver supervision.
In this subsection, we first present the existence of this uncertainty to explain the motivation behind tag-based label enhancement, the method we propose to mitigate such uncertainty in the routing function training.
We calculate the entropy of rewards from QwenRM among all candidate LLMs for each query in MT-Bench and draw it with the MT-Bench scores of each sample by reward preference ranking with QwenRM.
As shown in \Cref{fig:rm_uncertainty}, samples with lower reward entropy tend to have high MT-bench scores.
We interpret this observation as higher reward entropy reveals more uncertainty in the reward.
Therefore, we propose tag-based label enhancement to leverage a tag-based prior to adjust reward entropy.

\xhdr{Label Enhancement}
\begin{table}[t]
    \centering
    \small
    \setlength{\tabcolsep}{0.5mm}{
    \begin{tabular}{cccccc}
    \toprule
    $\mathbf{\beta}$  & \textbf{AlpacaEval} & \textbf{FLASK} & \textbf{MT-Bench} & \textbf{Benchmarks} & \textbf{All} \\
    \midrule
    0	& 1.4 & 2.2 & 2.25 & 3.67 & 2.06 \\
    0.1	& 1.2 & 2.1	& 2.38 & 3.67 & 2.00 \\
    \rowcolor{Ocean} 0.3	& 1.2 & 1.9	& 2.50 & 3.67 & 1.97 \\
    0.5	& 1.2 & 2.2	& 3.12 & 3.67 & 2.23 \\ 
    0.7	& 1.2 & 2.2	& 3.38 & 4.00 & 2.31 \\
    0.9	& 1.2 & 2.3	& 3.12 & 4.00 & 2.31 \\
    1.0	& 1.2 & 2.3	& 3.25 & 4.00 & 2.34 \\
    \bottomrule
    \end{tabular}}
    \caption{Mean task rank~(MTR) of different $\beta$ in tag-based label enhancement across all benchmarks. The best value of $\beta$ is marked in blue.}
    \label{tab:label_enhancement_analysis}
\end{table}
The tag-based label enhancement proposed in \Cref{sec:zooter} contains a hyper-parameter $\beta$, which represents the trade-off between fine-grained sample-level rewards and coarse-grained tag-level rewards.
We conduct experiments to tune this hyperparameter and analyze how rewards in different granularities may influence the training of our routing function.
As shown in \Cref{tab:label_enhancement_analysis}, \modelname achieves the best performance when $\beta$ equals $0.3$, proving a combination of sample-level and tag-level rewards will benefit the reward distillation.
The ablation also shows the necessity of tag-based label enhancement.
Furthermore, distilling tag-level rewards~($\beta=0$) shows significantly better performance than distilling sample-level rewards~($\beta=1$), supporting the analysis that noises from the uncertainty of RMs in sample-level rewards damage reward distillation.




\section{Conclusion}
In this work, we revisit the complementary potential of open-source LLMs and reward model ranking of multiple off-the-shelf reward models, providing evidence to the effectiveness of LLM ensemble.
We propose \modelname, an efficient reward-guided routing method for ensemble off-the-shelf LLMs.
Comprehensive evaluation shows \modelname can outperform the best single model on average and even ensemble models by reward model ranking with significantly fewer computation overhead.
Valuable future works include diving deep into the interpretation of latent expertise in each LLM.



\bibliography{anthology,custom}
\bibliographystyle{acl_natbib}

\newpage
\appendix

\section{Datasets}\label{app:dataset}
\datasetname is a diverse mix instruction set from multiple open-source datasets with careful decontamination on all benchmarks evaluated in this work.
The detailed composition of \datasetname is report in \Cref{tab:dataset}.

\begin{table}[t]
    \centering
    \begin{tabular}{lc}
    \toprule
    \textbf{Dataset}  & \textbf{Amount} \\
    \midrule
    ultrachat & 18,588\\
    sharedgpt & 10432 \\
    wizardlm(sharedgpt) & 5325 \\
    wizardlm(alpaca) & 5145 \\
    alpaca & 2186 \\
    repair & 1034 \\
    openchat & 1033 \\
    flan & 862 \\
    math & 849 \\
    unnatural & 582 \\
    dmcc & 573 \\
    dolly & 560 \\
    oasst & 183 \\
    lima & 70 \\
    mbpp & 43 \\
    \bottomrule
    \end{tabular}
    \caption{Composition of \datasetname}
    \label{tab:dataset}
\end{table}

\end{document}